\documentclass{article}

\usepackage[utf8]{inputenc}
\usepackage[T1]{fontenc}
\usepackage{caption}
\usepackage{subcaption}
\usepackage{framed}
\usepackage{hyperref}

\usepackage{textcomp}
\usepackage{graphics}
\usepackage{graphicx}
\usepackage{color}
\usepackage[normalem]{ulem}

\usepackage{microtype}
\usepackage{booktabs}

\usepackage{amsmath,amsthm,amsxtra,amssymb}
\usepackage{amsfonts}

\usepackage[accepted]{icml2019}

\DeclareMathOperator*{\argmin}{arg\,min}

\begin{document}

\twocolumn[
\icmltitle{Information-Bottleneck Approach to Salient Region Discovery}

\begin{icmlauthorlist}
\icmlauthor{Andrey Zhmoginov}{goog}
\icmlauthor{Ian Fischer}{goog}
\icmlauthor{Mark Sandler}{goog}
\end{icmlauthorlist}

\icmlaffiliation{goog}{Google Inc.}

\icmlcorrespondingauthor{Andrey Zhmoginov}{azhmogin@google.com}

\icmlkeywords{Machine Learning, ICML}

\vskip 0.3in
]

\printAffiliationsAndNotice{}  

\begin{abstract}
	We propose a new method for learning image attention masks in a semi-supervised setting based on the Information Bottleneck principle.
	Provided with a set of labeled images, the mask generation model is minimizing mutual information between the input and the masked image while maximizing the mutual information between the same masked image and the image label.
	In contrast with other approaches, our attention model produces a Boolean rather than a continuous mask, entirely concealing the information in masked-out pixels.
	Using a set of synthetic datasets based on MNIST and CIFAR10 and the SVHN datasets, we demonstrate that our method can successfully attend to features known to define the image class.
\end{abstract}

\def\npar{\vskip .3cm \noindent}
\def\I{\mathbb{I}}
\def\H{\mathbb{H}}
\def\E{\mathbb{E}}
\def\R{\mathbb{R}}
\newcommand{\newthought}[1]{#1.}
\newcommand{\alert}[1]{{\bf \color{blue} #1}}
\newcommand{\remove}[1]{{\color[gray]{0.6} #1}}
\newcommand{\Eq}[1]{Eq.~\eqref{#1}}

\def\oC{\mathrm{C}}
\def\oR{\mathrm{Resize}}
\def\oT{\mathrm{T}}
\def\oP{\mathrm{Pad}}
\def\oS{\mathrm{Shape}}
\def\oA{\mathrm{Avg}}
\def\oFC{\mathrm{FC}}

\section{Introduction}

    Information processing in deep neural networks is carried out in multiple stages and the data-processing inequality implies that the information content of the input signal decays as it undergoes consecutive transformations.
    Even though this applies to both information that is relevant and irrelevant for the task at hand, in a well-trained model, most of the useful information in the signal will be preserved up to the network output.
    However, standard objectives, such as the cross-entropy loss, do not constrain the irrelevant information that is retained in the output.

    The Information Bottleneck (IB) framework \cite{tishby2000information,tishby2015deep} constrains the information content retained at the output by trading off between prediction and compression: $IB \equiv \min \beta \I(X;Z) - \I(Y;Z)$, where $X$ is the input, $Y$ is the target output, and $Z$ is the learned representation.
    This framework has been applied to numerous deep learning tasks including a search of compressed input representations \cite{alemi2016deep,hjelm2018,moyer2018}, image segmentation \cite{bardera2009segmentation}, data clustering \cite{strouse2019,still2003}, generalized dropout \cite{achille2018dropout}, Generative Adversarial Networks \cite{peng2018} and others.
    
    In this paper, we use the IB approach to generate self-attention maps for image classification models, directing model attention away from distracting features and towards features that define the image label.
    The method is based on the observation that the information content of the image region that we want to ``attend to'' should ideally be minimized while still being descriptive of the image class.

    The proposed technique can be thought of as a form of semi-supervised attention learning.
    The entire model consisting of the mask generator and the classifier operating on the masked regions can also be viewed as a step towards ``explainable models'', which not only make predictions, but also assign importance to particular input components.
    This technique could potentially be useful for datasets that cannot be easily annotated by experts, such as medical image datasets where labels are known, but the particular cause of the label in the input is difficult to collect.

    The paper is structured as follows.
    Section~\ref{sec:prior} describes prior work and relates our approach to other existing methods.
    In Section~\ref{sec:theory} we outline theoretical foundations of our method and the experimental results are summarized in Section~\ref{sec:experiments}.
    Section~\ref{sec:extensions} discusses an alternative IB-based approach and finally, Section~\ref{sec:conclusions} summarizes our conclusions.

\section{Prior Work}
\label{sec:prior}

    Semi-supervised image segmentation is a task of learning to identify object boundaries without access to the boundary groundtruth information.
    Object detection and reconstruction of object shape in the context of this task is frequently achieved based on the knowledge of image labels alone \cite{hou2018,wei2017,zhang2018adversarial,li2018tell,kolesnikov2016}.
    A successfully trained model would thus effectively ``know'' which parts of the input image carry information defining the image class and which parts are irrelevant.
    Most of these methods use (in one or another way) a signal supplied by the classification model with a partially occluded input.
    By changing the attention mask and probing classifier performance it is possible to identify ``salient'' regions, as well as those regions that are not predictive of the object present in the image.

    Most semi-supervised semantic segmentation approaches including those mentioned above tend to rely on hand-designed optimization objectives and supplementary techniques that are carefully tuned to work well in specialized domains.
    In contrast, more general frameworks, like those based on information theory, could provide a more elegant and universal alternative.
    In recent years, the Information Bottleneck method has been applied to generating instance-based input attention maps.
    Most notably, an information-theoretic generalization of dropout called Information Dropout \cite{achille2018dropout} based on element-wise tensor masking was shown to successfully generate representations insensitive to nuisance factors present in the model input.
    Another novel approach called InfoMask recently proposed in \citet{taghanaki2019} independently of our work, applies IB-inspired approach to generating continuous attention masks for the image classification task.
    The authors demonstrated superior performance of InfoMask on the Chest Disease localization task compared to multiple other existing methods.
    
    In this work, we propose an alternative approach using the Information Bottleneck optimization objective.
    In contrast to two described approaches, we target the information content of the masked image and we do not multiply image pixels by a floating-point continuous mask, but instead use Boolean masks, thus completely preventing masked-out pixels from propagating any information to the model output.
    
\section{Model}
\label{sec:theory}

    Consider a conventional image classification task trained on samples drawn from a joint distribution $p(I,C)$ with the random variable $I$ corresponding to images and $C$ being image classes.
    Let us tackle a complimentary task of learning a self-attention model that given an image $i$ produces such a Boolean mask $m_\zeta(i)$ that the masked image $i\odot m_\zeta(i)$ satisfies two following properties: (a) it captures as little information from the original image as possible, but (b) it contains enough information about the contents of the image for the model to predict the image class.

    Using the language of information theory these two conditions can be satisfied by writing a single optimization objective:
    \begin{gather}
        \label{eq:orig_ib_pre}
        \min_\zeta Q_\beta \equiv \min_\zeta \left[ \beta \I(I\odot M;I) - \I(I\odot M;C)\right],
    \end{gather}
    where $\beta$ is a constant and $M$ is a random mask variable governed by some learnable conditional distribution $p_\zeta(m|i)$.
    Being written in the form of \Eq{eq:orig_ib_pre}, our task can be seen as a reformulation of the Information Bottleneck principle \cite{tishby2000information}.
    Alternative optimization objectives based on, for example, Deterministic Information Bottleneck \cite{strouse2017deterministic} could also be of interest, but fall outside of the scope of this paper.
    Another optimization objective based on the Conditional Entropy Bottleneck~\citep{ceb} is discussed in Appendix~\ref{sec:ceb}.

    Notice that Equation~\eqref{eq:orig_ib_pre} has one significant limitation: it allows the masking model to deduce the class from the image and encode this class in the mask itself.\footnote{Assuming it is sufficiently complex and has a receptive field covering the entire image.}
    Consider a binary classification task.
    If the image belongs to the first class, the generated mask can be empty.
    On the other hand, for images belonging to the second class, the mask can be chosen to be just a single or a few pixels taken from the ``low entropy'' part of the image thus both minimizing $\I(I\odot M;I)$ and maximizing $\I(I\odot M;C)$.
    For this choice of mask, the classifier $f_\psi$ can predict the label just from the mask itself.

    This unwanted behavior can be avoided in practice by choosing mask models with a finite receptive field that is comparable to the size of the feature distinguishing one class from another.
    A more general approach has to rely on special properties of the mask $m$.
	One such defining property is that $\I(I\odot M;C) \le \I(I\odot M';C)$ for Boolean masks $M'$ ``larger'' than $M$ in a sense that $m'_{x,y}=0$ implies that $m_{x,y}=0$.
    We can define $M'$ by, for example, specifying $p(m'_{x,y}|m_{x,y},x,y)$ and restricting it via $p(m'_{x,y}=0|m_{x,y}=1)=0$.
    Defined like this, our optimization objective \eqref{eq:orig_ib_pre} can be rewritten as:
    \begin{gather}
        \label{eq:orig_ib}
        \min_\zeta Q_\beta \equiv \min_\zeta \left[ \beta \I(I\odot M;I) - \I(I\odot M';C)\right].
    \end{gather}
    Preliminary exploration of the effect that {\em mask randomization} technique has on attention regions is presented in Section~\ref{sec:random}.

    
\subsection{Variational Upper Bound}

    Expanding the expressions for the mutual information in \Eq{eq:orig_ib}, we obtain:
    \begin{multline}
        \label{eq:q_beta}
        Q_\beta = \beta \H(I\odot M) - \beta \H(I\odot M|I) - \\ - \H(C) + \H(C|I\odot M').
    \end{multline}
    Entropies of the form $\H(A)$ permit variational upper bounds of the form $-\E_{a} \log p_\phi(a)$ with $p_\phi(a)$ taken from an arbitrary family of distribution functions, and similarly for conditional entropies $\H(A|B)$.
    This allows us to formulate the variational optimization objective as \cite{alemi2016deep}:
    \begin{multline}
        \label{eq:prefin_ib}
        \min_{\zeta,\theta,\psi} \biggl[ \E_{p(i,c)p_\zeta(m|i)} \biggl( -\beta \log g_\theta(i\odot m) - \\ - \log h_\psi(c|i\odot m') \biggr) - \beta \H(I\odot M|I) \biggr],
    \end{multline}
    where $g_\theta$ and $h_\psi$ are variational approximations of $p(i\odot m)$ and $p(c|i\odot m')$ correspondingly.
    Below, we compute $\H(I\odot M|I)$ explicitly for our choice of mask model.

\subsection{Mask and Masked Image}
    Let $\rho_\zeta: X \to \R^{n\times n}$ be the ``masking probability'' model parameterized\footnote{We will frequently be omitting $\zeta$ for brevity.} by $\zeta$.
    Each $\rho_{x,y}(i)$ for $1\le x,y \le n$ is assumed to satisfy $0\le \rho_{x,y}(i) \le 1$.
    We introduce a discrete mask $m=\textrm{Bernoulli}(\rho)$ sampled according to $\rho$ independently for each pixel.
    The masked image $i\odot m$ can then be defined as follows:
    \begin{gather}
        \label{eq:discmask}
        (i\odot m)_{x,y} \equiv
        \begin{cases}
            (i_{x,y}, 1) & \quad \textrm{if\,} m_{x,y}=1, \\
            (0, 0) & \quad \textrm{if\,} m_{x,y}=0.
        \end{cases}
    \end{gather}
    Given this definition, the entropy $\H(I\odot M|I)$ can be expressed as:
    \begin{gather}
        \label{eq:cont}
        - \sum_{x,y=1}^{n} \left[ \rho_{x,y} \log \rho_{x,y} + (1 - \rho_{x,y}) \log (1 - \rho_{x,y}) \right].
    \end{gather}

    It is worth noticing here that the mask $\rho_\zeta(i)$ can be interpreted as an adaptive ``continuous'' downsampling of the image.
    Low values of $\rho$ cause most, but not all image pixels to be removed; the remaining pixels and the mere fact that the mask chose to partially remove them can still provide enough information to the image classification model.

\subsection{Loss Function}
    Having the expression for the last term in \Eq{eq:prefin_ib}, we will now provide specific models for the first two.

    Let us start with $-\log h_\psi(c|i\odot m')$.
    Consider a family of deep neural network models $f_\psi$ mapping masked images $i\odot m'$ to $\R^{|c|}$.
    We can define $h_\psi(i\odot m')$ to be $\textrm{softmax}\,(f_\psi(i\odot m'))$ allowing us to rewrite $-\log h_\psi(c|i\odot m')$ as a cross-entropy loss with respect to $\textrm{softmax}\,(f_\psi(i\odot m'))$.
    Recalling that mask $m$ is sampled from $\textrm{Bernoulli}(\rho_\zeta)$, we cannot simply back-propagate gradients all the way down to the parameters of the model $\rho_\zeta(i)$.
    We alleviate this problem by using the Gumbel-softmax reparametrization approach~\citep{jang2016categorical,maddison2016concrete},
    \footnote{It is worth mentioning that the Gumbel temperature should be chosen with care; very small values lead to high-variance estimators, while low temperature would introduce bias.}
    thus approximating $m(i)$ with a differentiable function.

\begin{figure}  
        \centering
        \includegraphics[width=0.48\textwidth]{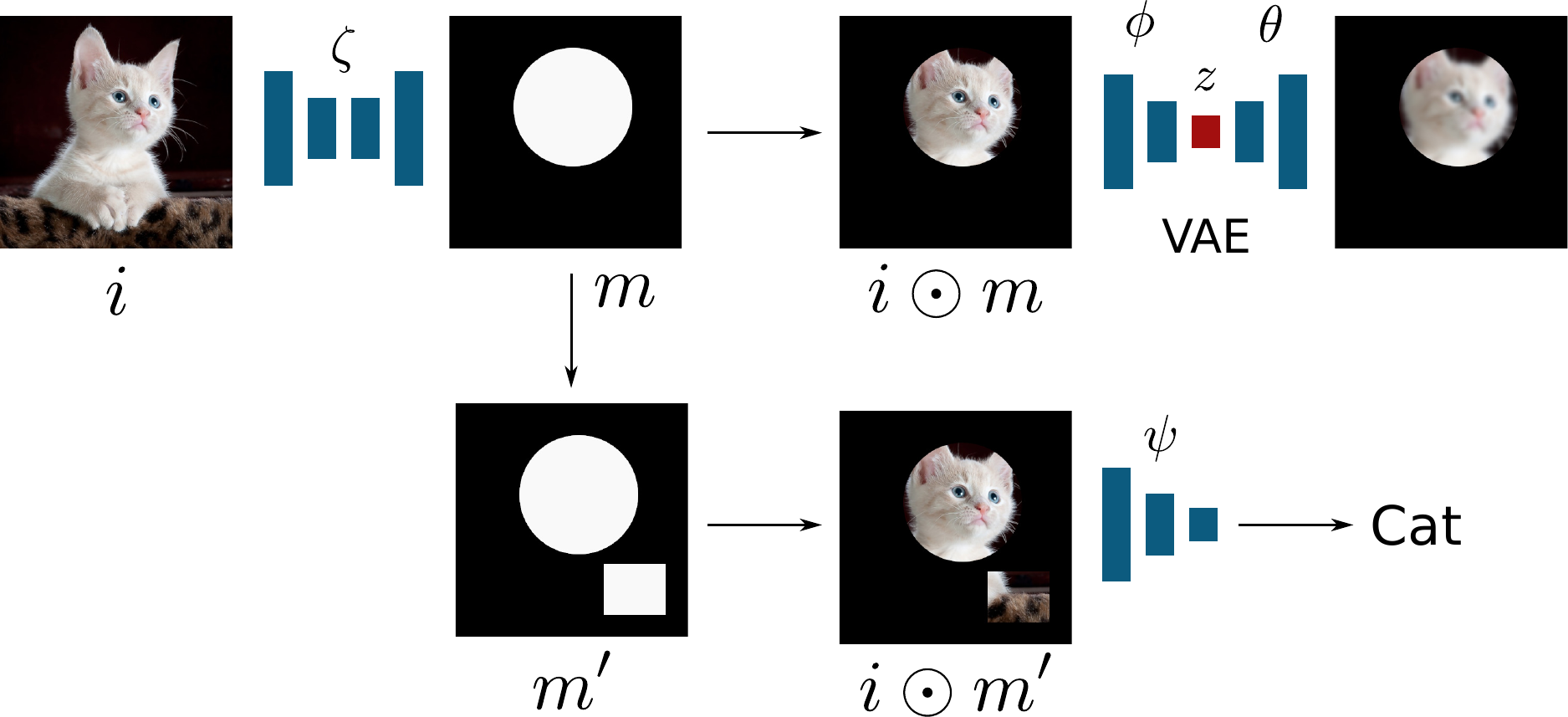}
        \caption{Model diagram: (a) the image $i$ is used to produce masking probability $\rho_\zeta(i)$ and the mask $m$ is then sampled from $\textrm{Bernoulli}(\rho_\zeta(i))$, (c) the mask $m$ is randomly augmented (grown) to produce $m'$, (d) the original masked image $i\odot m$ is autoencoded via $(g_\theta, q_\phi)$, (e) the masked image $i\odot m'$ is used as an input to a classification model $f_\psi:i\odot m'\mapsto c$.}
        \label{fig:diagram}
    \end{figure}

    Now let us consider the first term in \Eq{eq:prefin_ib}.
    Since the space of masked images $i\odot m$ is generally very high-dimensional, we adapt the variational autoencoder approach \cite{kingma2013auto}, considering a space of marginal distribution functions $g_\theta(i\odot m)=g_\theta(i\odot m|z) p(z)$ with $p(z)$ being a tractable prior distribution for the latent variable space $Z$. 
    Following \citet{kingma2013auto}, $-\log g_\theta(i\odot m)$ can be upper bounded by:
    \begin{multline*}
        -\E_{z\sim q_\phi(z|i\odot m)} \left[ \log g_\theta(i\odot m|z) \right] + \\ + D_{\rm KL} \left[ q_\phi (z|i\odot m) \| p(z) \right],
    \end{multline*}
    where $q_\phi$ is a variational approximation of $g_\theta(z|i\odot m)$.
    The encoder $q_\phi$ in our model receives both the input pixels $i_{x,y}$ (or $0$ if $m_{x,y}=0$) and the mask $m_{x,y}$ as its inputs and produces a conventional embedding $z\in \R^d$.
    The decoder $g_\theta$, in turn, maps $z$ back to $\hat{\rho}$ and $\hat{i}$.
    In our model, we define $g_\theta(i\odot m|z)$ as a probability for a masked image to be sampled from a Bernoulli process with a probability $\hat{\rho}$ and the image to be sampled from a Gaussian random variable with the mean $\hat{i}$ and a constant covariance matrix.
    This allows us to rewrite $-\log g_\theta(i\odot m|z)$ as:
    \begin{multline}
        \label{eq:g}
        -\log g_\theta(i\odot m|z) = \sum_{x,y=1}^{n} \biggl\{ - (1 - m_{x,y}) \log (1 - \hat{\rho}_{x,y})
        - \\ - m_{x,y} \left[ \log \hat{\rho}_{x,y} - \ell_2(i_{x,y},\hat{i}_{x,y}) \right] \biggr\} + C,
    \end{multline}
    where $\ell_2(i,\hat{i}) = {(i - \hat{i})^2}/{2\sigma^2}$ and $\sigma$, $C$ are constants.
    Given this choice, $\beta$ becomes an overall multiplier of the VAE objective in the full loss and $\sigma$ defines a weight of the image pixel reconstruction relative to the mask reconstruction.
    The entire model is illustrated in Figure~\ref{fig:diagram}.

    It is worth noticing that adopting the Gumbel-softmax trick we find that a discrete approximation of \Eq{eq:cont} reading
    \begin{gather}
        \label{eq:discrete}
        - \sum_{x,y=1}^{n} \left[ m_{x,y} \log \rho_{x,y} + (1 - m_{x,y}) \log (1 - \rho_{x,y}) \right]
    \end{gather}
    leads to better convergence in our experiments.
    We hypothesize that better empirical performance of models using \Eq{eq:discrete} rather than \Eq{eq:cont} can potentially be explained by the fact that the Gumbel-softmax reparametrization introduces bias and therefore, expression in \Eq{eq:cont} will not cancel on average with the corresponding term \eqref{eq:g} in VAE even for the perfect mask reconstruction, i.e., $\hat{\rho}=\rho$.

\section{Experimental Results}
\label{sec:experiments}

    All our experiments were conducted for the original optimization objective \eqref{eq:orig_ib} by optimizing the loss function derived in Section~\ref{sec:theory} using \Eq{eq:discrete} instead of \Eq{eq:cont}.
    We observed that the behaviour of the model was very sensitive to the constant $\beta$.
    If $\beta$ was too small, the mask $\rho_\zeta$ would monotonically approach $\rho_\zeta=1$.
    Conversely, for sufficiently large $\beta$, $\rho_\zeta$ would vanish.
    We used two different techniques to improve behaviour of our model: (i) stop masking model gradients in variational autoencoders once $-\log g_\theta$ falls below a certain threshold, or (ii) change $\beta$ adaptively in such a way that $-\log g_\theta$ stays within a pre-defined range.
    Both of these approaches were able to guarantee in practice that the variational autoencoder loss reached a certain predefined value.
    For additional details of our model, see Appendix~\ref{sec:details}.

    In all experiments discussed in this section, the groundtruth ``features'' that define image class are known in advance allowing us to interpret experimental results with ease.
    In a more general case, the quality of the model prediction can be judged based on the following three criteria: (a) accuracy of the trained classifier operating on masked images $I\odot M$ should be sufficiently close to the accuracy of a separate classifier trained on original images $I$; (b) VAE loss should fall into a predefined range; (c) the accuracy of the classifier prediction on $I\odot M'$ should be sufficiently close to the prediction on $I\odot M$ for any fixed $I$ and all sampled realizations of $M'$.

    In the following subsections, we first discuss our results on synthetic datasets with ``anomalies'' and ``distractors''.
    These experiments were conducted without mask randomization, but we verified that experiments with mask randomization produced nearly identical results.
    We then discuss our experiments on a synthetic dataset designed to explore the effect that mask randomization has on produced masks.
    Finally, we show results on a realistic {\tt SVHN} dataset with apriori known localized features defining the image class (number of digits in the image).
    For this dataset, mask randomization appears to play an important role.

\subsection{Experiments with ``Anomalies''}

    For the first series of experiments, we used images from {\tt CIFAR10} \cite{krizhevsky2009learning} and {\tt MNIST} datasets augmented by adding randomly-placed rectangular ``anomalies'' (thus designed to be low-entropy).
    The anomaly was added with a probability of $1/2$ and the classification task was to distinguish original images from the altered ones.

\begin{figure} 
        \centering
        \includegraphics[width=0.4\textwidth]{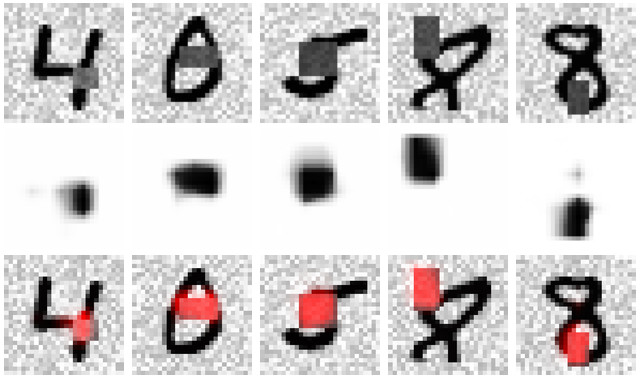}
        \caption{Results for the {\tt MNIST} dataset with rectangular patches: augmented images (top row); masks (middle row; white represents opaque regions, black transparent); mask on top of the augmented image (bottom row).}
        \label{fig:mnist_rect}
    \end{figure}

\begin{figure}[htpb]
        \centering
        \includegraphics[width=0.4\textwidth]{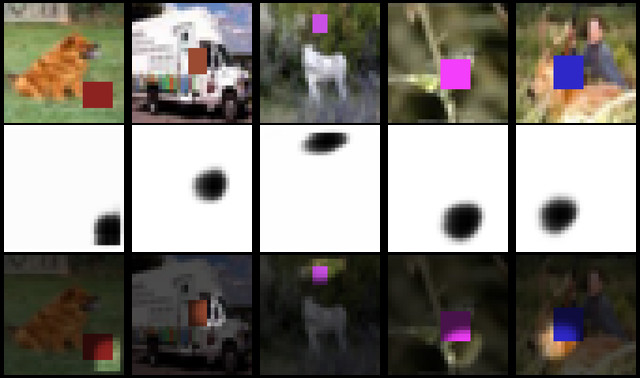}
        \caption{Results for the {\tt CIFAR10} dataset with rectangular patches: augmented images (top row); masks (middle); mask on top of the augmented image (bottom).}
        \label{fig:cifar_rect}
    \end{figure}

    For these datasets, our models learned to produce opaque masks for most images without anomalies.
    For images with anomalies, generated masks were opaque everywhere except for the regions around rectangles added into the image (see Figure~\ref{fig:mnist_rect} and Figure~\ref{fig:cifar_rect}).
    As a result, the image classifiers reached almost perfect accuracy in both of these examples: approximately $98\%$ test and train accuracy for {\tt MNIST} and approximately $99\%$ test and train accuracy for {\tt CIFAR10} dataset.

    In both models, $\ell_1$ norm of the mask was a strong predictor of whether the ``anomaly'' was in the image (see Figures~\ref{fig:mnist_stats} and \ref{fig:cifar_stats}).
    However, interestingly, the separation was much more visible for {\tt CIFAR10}, while the masks predicted for the {\tt MNIST} dataset were much better aligned with the actual anomalies.
    The latter fact can also be seen to be reflected in the mask averages inside and outside of the actual added rectangles (see Figures~\ref{fig:mnist_stats} and \ref{fig:cifar_stats}).

    We hypothesize that these properties of the trained models can be attributed to receptive fields of the masking models used in both examples.
    For the {\tt MNIST} dataset, the masking model has a receptive field of about 40\% of the image size, while for the {\tt CIFAR10} dataset, the receptive field covered nearly the entire input image.

 \subsection{Experiments with ``Distractors''}

    In another set of experiments, we used two synthetic datasets based on {\tt MNIST}, in which we combined: (a) two digits and (b) four digits in a single $56\times 56$ image.
    In both datasets, one of the digits was always smaller and it defined the class of the entire image.
    The larger digits are thus ``distractors''.

\begin{figure} 
        \centering
        \includegraphics[width=0.45\textwidth]{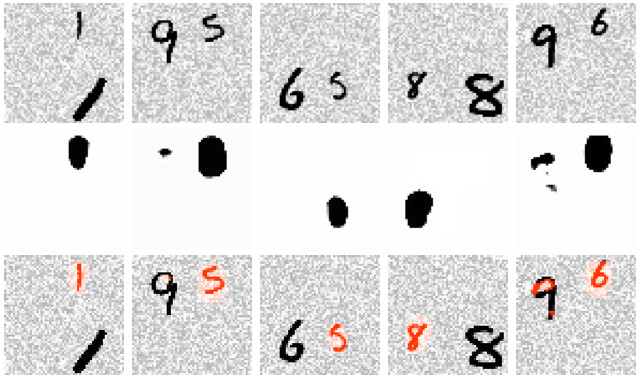}
        \caption{Results for the double-digit {\tt MNIST}-based dataset: original images (top row); learned masks (middle); mask on top of the original image (bottom). Images on the right demonstrate one of the failures of the model.}
        \label{fig:double_mnist}
    \end{figure}

\begin{figure} 
        \centering
        \includegraphics[width=0.45\textwidth]{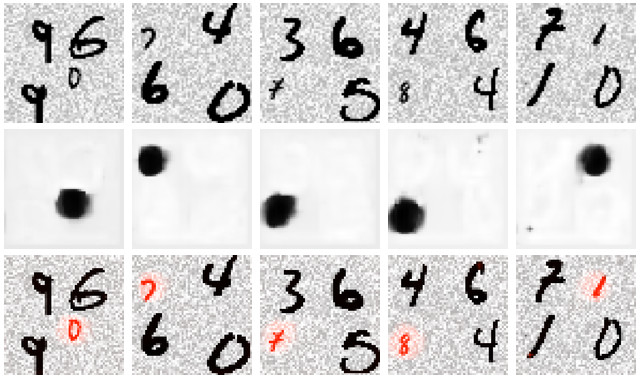}
        \caption{Same as Figure~\ref{fig:double_mnist}, but for the four-digit {\tt MNIST}-based dataset.}
        \label{fig:quad_mnist}
    \end{figure}

    For the vast majority of masks generated by the trained model, everything outside of the region around the small digit was masked-out (see Figures~\ref{fig:double_mnist}, \ref{fig:quad_mnist} and \ref{fig:quad_stats}).
    In some rare cases, however, generated masks were also letting some pixels of the larger digits to pass through.
    In most of our experiments, the classifier training and test accuracy reached $95\%$ and $90\%$ for the two- and four-digits datasets correspondingly.
    However, there were some runs, in which the test accuracy could be lower than the training accuracy by $10\%$ or $20\%$.
    We believe this is due to the greater capacity to overfit to the training data of the combined masking and classifier models.
 
\subsection{Mask Randomization Experiments}
\label{sec:random}

\begin{figure*} 
        \centering
        \begin{subfigure}{0.75\textwidth}
          \centering
          \includegraphics[width=.99\linewidth]{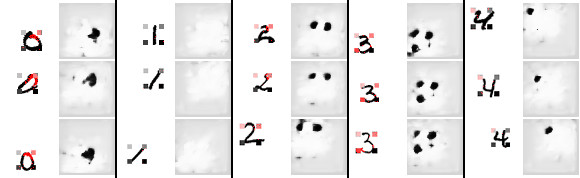}
          \caption{}
          \label{fig:no_randomization}
        \end{subfigure}
        \begin{subfigure}{0.75\textwidth}
          \centering
          \includegraphics[width=.99\linewidth]{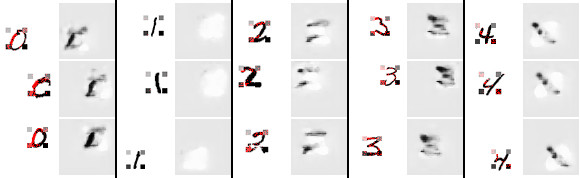}
          \caption{}
          \label{fig:with_randomization}
        \end{subfigure}
        \caption{Masked images and corresponding masks for models without (a) and with (b) {\em mask randomization}. Red shows pixels visible to the classifier and black pixels are masked-out by the learned attention region. Masks use white color to show pixels that are most likely to be masked-out and black for pixels that are most likely to be visible to the classifier. (a) Results for a model trained without mask randomization. Notice that the model rarely chooses actual digit pixels and is frequently seen to encode digit class into the ``anchors''; (b) Results for a model trained with mask randomization. Now the attention is directed towards digit pixels.}
        \label{fig:mnist_m}
    \end{figure*}

    We identified a simple MNIST-based synthetic example, in which it can be clearly seen that without mask randomization, generated masks can encode class information without using virtually any pixels from the actual digits.
    In our example, we use 5 MNIST digits (0 through 4) and add 4 solid rectangles (``{\em anchors}'') into the image thus allowing the mask to use them for encoding image label.
    Model trained without any mask randomization, i.e., $M'=M$ can be seen to produce attention regions selecting anchors, but frequently avoiding actual digit pixels altogether (see Figure~\ref{fig:no_randomization}).
    Trained classifier has almost perfect accuracy ($\sim 99\%$) on original masked images.
    However, once we start evaluating the same classifier on images with randomized masks (adding random transparent rectangular patches), the accuracy drops down to $\sim 33\%$ for some of the digits.
    After the classifier is fine-tuned on images with randomized masks, the lowest accuracy for a digit goes up to $70.3\%$ (for digit $2$, which ends up being most frequently confused for $3$).

    We then conduct experiments with the same dataset and enable mask randomization during training (by selecting $M'$ to be equal to $M$ with a randomly placed transparent rectangle).
    New trained models now mainly concentrate on the digit pixels and seem to select discriminative parts of the image (see Figure~\ref{fig:with_randomization}).
    Evaluating the accuracy of this classifier with mask randomization, we observe that the average accuracy now stays above $93\%$ for all digits.

\subsection{{\tt SVHN} Experiments}
\label{sec:svhn_exp}

    We chose the original {\tt SVHN} dataset \cite{netzer2011reading} for our experiments with realistic images.
    The task given to a classifier was to predict the number of digits in the street/house number shown in the image.
    With this task, the generated mask was expected to concentrate on areas of the image containing numbers.

\begin{figure*} 
        \centering
        \includegraphics[width=0.7\textwidth]{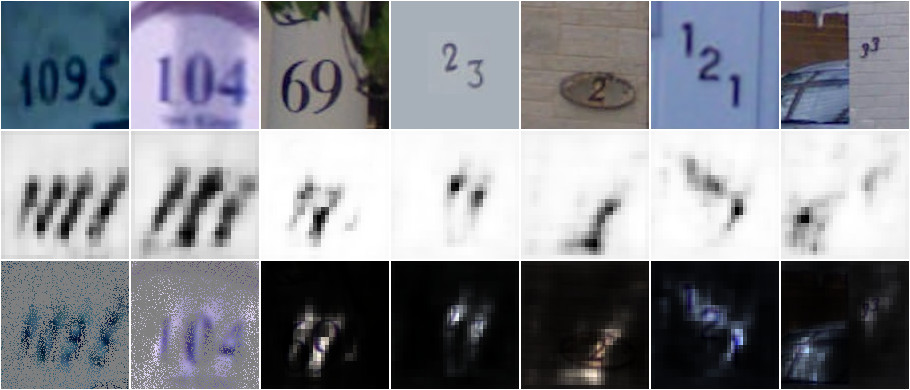}
        \caption{Attention results for {\tt SVHN} dataset: original image (top row); mask (middle row) and masked image (bottom row). First two columns show results for typical test images obtained using the same data augmentation procedure as the training images (masked images use Boolean mask for these images and are similar to masked images actually seen by the classifier during training); remaining columns show results on out-of-distribution samples obtained by cropping out $128\times 128$ regions from high-resolution source images.}
        \label{fig:svhn}
    \end{figure*}

    We started our experiments without mask randomization, i.e., $m'=m$.
	We picked $\sigma=(1/8)^{1/2}$ and the target VAE loss objective was chosen in such a way that the mask was neither transparent, nor almost completely opaque.
    For intermediate values of the VAE objective, most of the observed solutions produced noticeable peaks of transparency around the digits.
    Results obtained for one of the models trained with sufficiently low VAE target are shown in Figure~\ref{fig:svhn}.
    
    For lower VAE loss targets, we frequently observed masks that used interleaved transparent and opaque lines (either vertical or horizontal) as means of minimizing VAE loss while still allowing the classifier to achieve high accuracy in predicting the number of digits in the image.
    
\begin{figure} 
        \centering
        \includegraphics[width=0.5\textwidth]{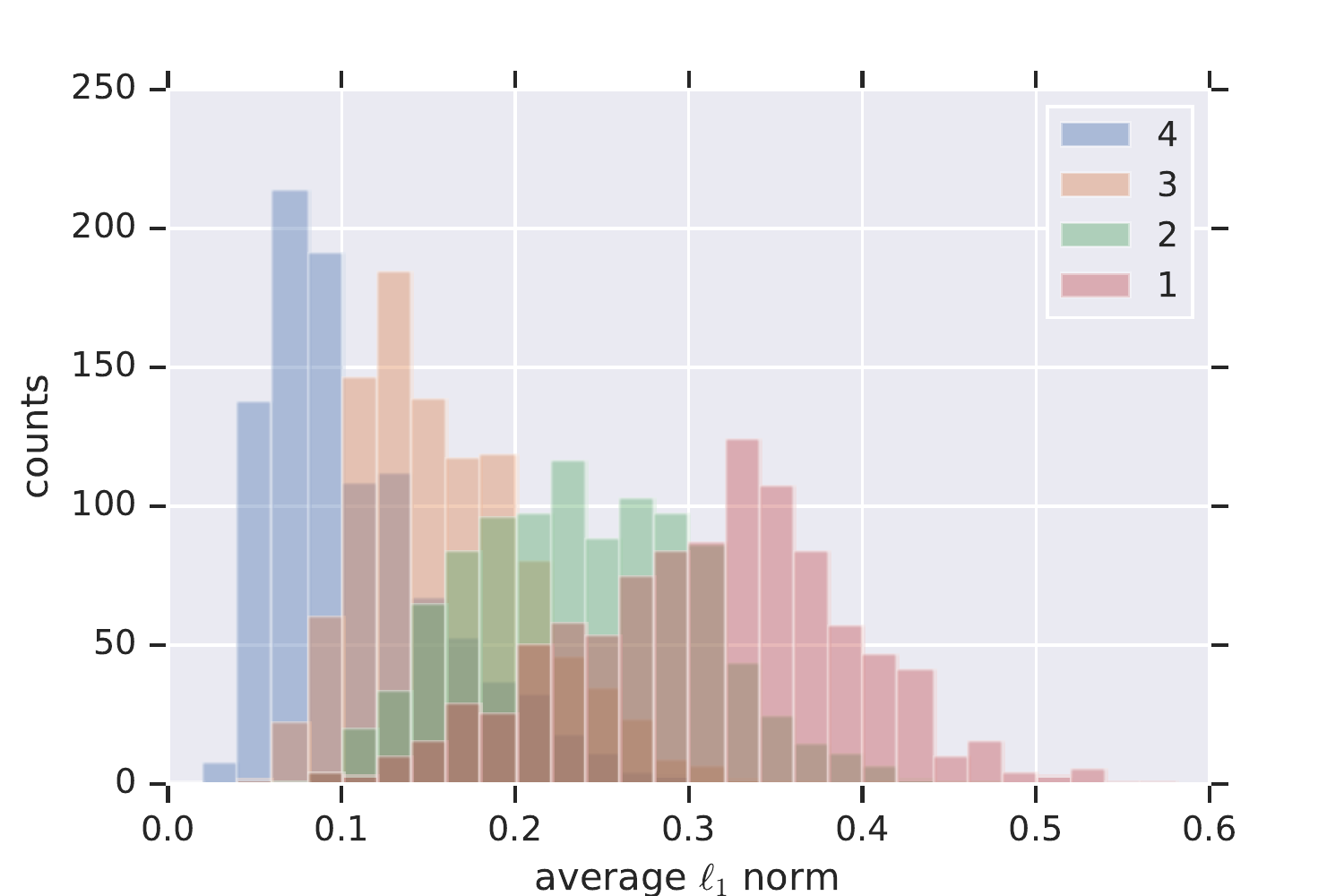}
        \caption{Average per-pixel $\ell_1$ mask norm for test samples with $1$, $2$, $3$ and $4$ digits.}
        \label{fig:svhn_hist}
    \end{figure}

    For even lower VAE thresholds, generated masks were no longer transparent around the digits, but instead were mostly opaque in these areas.
    This behavior can be understood by noticing that the digits containing many complex sharp edges may carry more information\footnote{also poorly approximated by VAEs, which tend to favor smooth reconstructions} than relatively featureless surrounding areas.
    In essence, the ``negative space'' outside of the number bounding box may be smaller-entropy, but its shape may still be enough to determine the number of digits in the image.
    In this case, the mask itself became a feature strongly correlated with the image label.
    Plotting histograms for $\ell_1$ mask norm, we observed that masks generated for $1$-digit images were almost entirely transparent while the masks generated for $4$-digit images were mostly opaque. 
    The histogram of $\ell_1$ mask norm for different image labels is shown in Figure~\ref{fig:svhn_hist}.
    We verified that using the $\ell_1$ mask norm alone, we could reach a $59\%$ accuracy on the image classification task, just $3\%$ lower than the actual trained classifier receiving the masked image.

    If all digits had the same aspect ratio, attending to the ``negative space'' of the number could actually be a reasonable solution satisfying all conditions outlined in Section~\ref{sec:theory}.
    In a more general case, observed masks that simply encode image class information do not seem to satisfy the condition $\I(I\odot M;C) \le \I(I\odot M';C)$.
    Implementing mask randomization by adding randomly-placed transparent rectangles to $m$, we verified that newly trained masking models were now nearly always concentrating on digits rather than the ``negative space''.

\section{Alternative Approach based on Conditional Mutual Information}
\label{sec:extensions}

    In previous sections, we showed that the Information Bottleneck optimization objective \eqref{eq:orig_ib_pre} allows for the class information to be encoded in the mask itself.
    Previously, we used mask randomization to address this issue.
    Another approach to disallowing the generated mask to encode class information is based on modifying the Information Bottleneck objective by replacing $\I(I\odot M;C)$ with $\I(I\odot M;C|M)$ thus leading to the optimization objective:
    \begin{gather}
        \label{eq:cond_ib}
        \argmin_\zeta \left[ \beta \I(I\odot M;I|M) - \I(I\odot M;C|M)\right],
    \end{gather}
    where we also chose to minimize $\I(I\odot M;I|M)$ instead of $\I(I\odot M;I)$ for consistency.
    Conditioning on the mask implies that for any realization of the mask, masked pixels should contain the entirety of the information about the image class.
    If, for example, the image class could be inferred just from the mask, the conditional mutual information $\I(I\odot M;C|M)$ would vanish.

    In order to optimize this objective, we have to modify \Eq{eq:cond_ib} by introducing a function $c'(i)$ that is chosen to approximate the groundtruth label $C$:
    \begin{gather}
        \label{eq:cond_ib_mod}
        \argmin_\zeta \left[ \beta \I(I\odot M;I|M) - \I(I\odot M;C'|M)\right].
    \end{gather}
    The exact form of $c'(i)$ will prove to be unimportant and in practice we frequently chose actual labels for our experiments assuming that the perfect groundtruth model $I\to C$ exists.

    As shown in Appendix~\ref{sec:condition_mask}, the optimization problem \eqref{eq:cond_ib_mod} is equivalent to:
    \begin{multline}
        \label{eq:cond_ib_exp}
        \argmin_\zeta \biggl[ \beta \H(I\odot M|M) + \H(M) - \H(M|I) + \\ + \H(I|M,C') + \H(C'|I\odot M)\biggr].
    \end{multline}
    Following our earlier discussion, we can then explicitly calculate $\H(M|I)$ and use variational upper bounds for all remaining entropies and conditional entropies.
    The complete model will therefore include: (a) VAE on the masked portion of the image $I\odot M$ conditioned on the image mask $M$; (b) VAE for the mask $M$ itself, (c) VAE auto-encoding the image $I$ and conditioned on the mask $M$ and the class approximation $C'$ and (d) image classifier with $I\odot M$ as its input.
    Notice that it is the conditional entropy $\H(I|M,C')$ that is responsible for disentangling $M$ and $C'$.
    Indeed, trying to minimize the entropy of images conditioned on the mask and the image class, we effectively reward the mask for containing information from $I$ that is not encoded in $C'$.

\begin{figure*} 
        \centering
        \includegraphics[width=.75\linewidth]{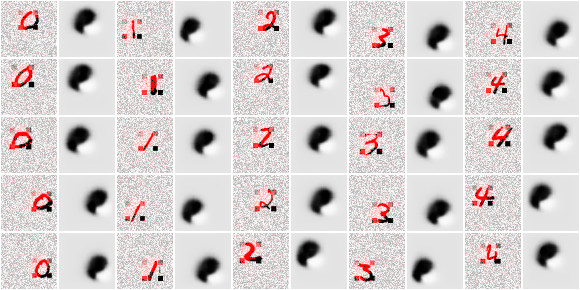}
        \caption{
            Same as Figure~\ref{fig:mnist_m}, but showing masked images and corresponding masks for the model based on \Eq{eq:cond_ib_exp} (in this dataset, we also added background noise to introduce a ``price'' of autoencoding background).
            Notice that the generated mask is almost indepent of the image class.
        }
        \label{fig:new_mask}
   \end{figure*}

    In our first preliminary experiments, we trained the upper-bound model for \eqref{eq:cond_ib_exp} on the MNIST-based synthetic datasets.
    For the dataset with ``anchors'', the model was able to generate masks that were: (a) covering the digits, (b) allowing the classification model to achieve $94\%$ accuracy and (c) nearly independent of the image label (see Figure~\ref{fig:new_mask}), which is exactly what objective \eqref{eq:cond_ib_exp} was designed to achieve.
    Similarly, for the dataset with distractors, the generated masks were almost indistinguishable from those shown in Figure~\ref{fig:quad_mnist}.
    
    For the dataset with anomalies, the experiments based on \Eq{eq:cond_ib_exp} failed to identify a proper mask and instead produced a mask transparent at the boundary and almost entirely opaque at the image center.
    Average masking probability $\langle \rho \rangle$ in the center encoded information about the image class so that the classifier could (with accuracy close to $100\%$) predict the presence of anomaly by just averaging values of visible pixels.
    This failure is not surprising if you notice that a mask transparent near an anomaly (see for example Figure~\ref{fig:mnist_rect}), but opaque for an image without one, does not optimize objective~\eqref{eq:cond_ib_exp}.
    Indeed, given such a mask, one would be able to predict image class by just looking at the mask itself.
    The optimal mask would have to have shape and location independent of the image class and concentrate on anomaly if it is present in the image.
    Instead of finding this complex solution, our model identified a simpler one by producing a mask that is {\em almost} independent of the image class, but still conveys enough information about the presence of anomaly in the image.

    Overall, while being conceptually sound, objective \Eq{eq:cond_ib_exp} is much more complex than \Eq{eq:q_beta} making it potentially less effective in practice.
    The disentangelement of $M$ and $C'$ critically relies on the upper bound for $\H(I|M,C')$ to be sufficiently tight and we suspect that it may be difficult to achieve this in practice for complex datasets containing realistic images.
    More complex density estimation models could, however, alleviate this problem.

\section{Conclusions}
\label{sec:conclusions}

    In this work, we propose a novel universal semi-supervised attention learning approach based on the Information Bottleneck method.
    Supplied with a set of labeled images, the model is trained to generate discrete attention masks that occlude irrelevant portions of the image, but leave enough information for the classifier to correctly predict the image class.
    Using synthetic and real datasets based on {\tt MNIST}, {\tt CIFAR10}, and {\tt SVHN}, we demonstrate that this technique can be used to identify image regions carrying information that defines the image class.
    In some special cases when the feature itself is high-entropy (for example, digits in {\tt SVHN} images), but its shape is sufficient to determine the image class (number of digits in our {\tt SVHN} example), we show that the generated mask may occlude the feature and use its ``negative space'' instead.
    Additionally, we identify a potential failure of this approach, in which the generated mask acts not as an attention map, but rather as an encoding of the image class itself.
    We then propose two techniques based on finite receptive fields and mask randomization that mitigate this problem.
    We believe this technique is a promising method to train explainable models in a semi-supervised manner.

\bibliography{bibliography}
\bibliographystyle{icml2019}

\appendix

\section{Relation to Conditional Entropy Bottleneck}
\label{sec:ceb}

    There is an alternative information-theoretic optimization objective that looks similar to \Eq{eq:orig_ib}, but is based on Conditional Entropy Bottleneck (CEB) \cite{ceb} instead:
    \begin{gather}
        \label{eq:orig}
        \min_\zeta Q'_{\beta'} \equiv \min_\zeta \left[ \beta' \I(I\odot M;I|C) - \I(I\odot M';C)\right].
    \end{gather}
    Here the mutual information between $I$ and $I\odot M$ is conditioned on the class label variable $C$.
    Just like $Q$, new objective $Q'$ can be rewritten as:
    \begin{multline*}
        Q'_{\beta'} = \beta' \left[ \H(I\odot M|C) - \H(I\odot M|I) \right] - \\ - \H(C) + \H(C|I\odot M').
    \end{multline*}
    Rewriting $\H(I\odot M|C)$ as $\H(C|I\odot M) + \H(I\odot M) - \H(C)$ and recalling that $\H(C)$ is a constant, $Q'$ can be expressed as:
    \begin{multline*}
        Q'_{\beta'} = \beta' \left[ \H(I\odot M) - \H(I\odot M|I) \right] + \\ + \H(C|I\odot M') + \beta' \H(C|I\odot M) + \nu.
    \end{multline*}
	where $\nu$ is a constant.
	Notice that without mask randomization when $M'=M$, this expression can be further rewritten as:
    \begin{gather*}
        Q'_{\beta'} = (1 + \beta') Q_{\beta'/(1+\beta')} + \nu'.
    \end{gather*}
    with $\nu'$ being a new constant.
    This suggests that for $M'=M$ the original optimization objective \eqref{eq:orig_ib} with $0 \le \beta < 1$ is equivalent to a CEB-based optimization objective \eqref{eq:orig} with $\beta' = \beta / (1 - \beta)$.

    By analogy with \Eq{eq:prefin_ib}, the upper bound for \Eq{eq:orig} can be written as:
    \begin{multline}
        \label{eq:prefin}
        \min_{\zeta,\theta,\psi} \biggl[ \E_{p(i,c)p_\zeta(m|i)} \biggl( -\beta' \log g_\theta(i\odot m|c) - \\ - \log h_\psi(c|i\odot m') \biggr) - \beta' \H(I\odot M|I) \biggr].
    \end{multline}
    The only difference from \Eq{eq:prefin_ib} is in the fact that the probabilistic model $g_\theta^{(c)}(i\odot m) = g_\theta(i\odot m|c)$ now depends on the sample class $c$.

    Even though in our experiments we used the original optimization objective \eqref{eq:orig_ib}, CEB-based objective \eqref{eq:orig} and the corresponding upper bound \eqref{eq:prefin} may have a practical advantage when the class-dependent variational approximations $g_\theta^{(c)}(i\odot m)$ provide a tighter bound than the class-agnostic model $g_\theta(i\odot m)$.

\section{Derivation of Equation~\eqref{eq:cond_ib_exp}}
\label{sec:condition_mask}

    First, following the definition of the conditional mutual information:
    \begin{multline*}
        \I(I\odot M;I|M) = \H(I\odot M|M) - \\ - \H(I\odot M|I,M) = \H(I\odot M|M).
    \end{multline*}
    Next, noticing that $\H(M,I)=\H(M,I,C')$, we obtain:
    \begin{gather*}
        \H(M|I) + \H(I) = \H(I|M,C') + \H(M,C')
    \end{gather*}
    and therefore
    \begin{multline*}
        \I(C';I\odot M|M) = \H(C'|M) - \H(C'|I\odot M,M) = \\ = \H(C',M) - \H(M) - \H(C'|I\odot M)
    \end{multline*}
    can be rewritten as:
    \begin{gather*}
        \H(M|I) + \H(I) - \H(I|M,C') - \H(M) - \H(C'|I\odot M).
    \end{gather*}
    Combining all terms together we obtain:
    \begin{multline*}
        \beta \I(I\odot M;I|M) - \I(C';I\odot M|M) = \\ = \beta \H(I\odot M|M) - \H(M|I) - \H(I) + \\ + \H(I|M,C') + \H(M) + \H(C'|I\odot M),
    \end{multline*}
    where $\H(I)$ is a constant.

\section{Model Details}
\label{sec:details}

    In our experiments, all of the model components including the classifier, mask generator and VAE encoder/decoder were based on convolutional neural networks.

\subsection{Notation}

    In the following, we use a simplified notation for writing down simple convolutional network architectures.
    Convolutional operation is denoted as $\oC(k,s,d)$ (default padding is {\tt valid}; subscript $s$ indicates {\tt same} padding), where $k$ is the kernel size, $s$ is the stride and $d$ is the number of output channels.
    Image resizing is denoted by $\oR(s)$ with $s$ being the new size and $\oS(s)$ is tensor reshaping.
    Similarly, $\oT(k,s,d)$ is the transpose convolution, $\oP(x)$ is image padding, $\oA$ is the average pooling operator and $\oFC(d)$ is the fully-connected layer mapping its input to a vector of size $d$.
    The architecture is then represented as a sequence of operations separated by $\rightarrow$, i.e., $a\rightarrow b=b\circ a$.

\subsection{{\tt MNIST}}

    For this dataset, we used original {\tt MNIST} images with randomly-placed rectangles.
    Rectangles were placed randomly and their size varied from 3 to 10 pixels.
    The color was randomly chosen from a range $[100,255]$.
    The VAE loss target was set at $12$.

    \def\RR{\textrm{ReLU6}}
    \def\LR{\textrm{Leaky\,ReLU}}

    \begin{itemize}
        \item {\bf Classifier architecture}: $\oC(1, 1, 4) \rightarrow \oC(3, 2, 4) \rightarrow \oC(1,1,8) \rightarrow \oC(3, 2, 8) \rightarrow \oC(1, 1, 16) \rightarrow \oC(3, 2, 16) \rightarrow \oA \rightarrow \oFC(2)$ with $\RR$ nonlinearities.
        \item {\bf Mask architecture}: $[\oC(1, 1, 4) \rightarrow \oC(3, 2, 4) \rightarrow \oC(1, 1, 8) \rightarrow \oC(3, 2, 8)] \rightarrow [\oR(12) \rightarrow \oC(1, 1, 16) \rightarrow \oP(1) \rightarrow \oR(28) \rightarrow \oC(1, 1, 16) \rightarrow \oC(1, 1, 1)]$ with the subnetwork in the first half using $\RR$ and the network in the second half using $\LR$.
        \item {\bf Encoder architecture}: $\oC(3, 2, 16) \rightarrow \oC(3, 2, 16) \rightarrow \oC(3,1,16)$ with $\LR$ nonlinearities.
        \item {\bf Decoder architecture}: $\oFC(24) \rightarrow \oFC(49) \rightarrow \oS(7\times 7) \rightarrow \oT_s(3, 2, 16) \rightarrow \oT_s(3, 1, 16) \rightarrow \oT_s(3, 2, 16) \rightarrow \oC(1, 1, 2)$ with $\LR$ nonlinearities.
    \end{itemize}

\subsection{{\tt CIFAR10}}

    For this dataset, we used original {\tt MNIST} images with randomly-placed rectangles.
    Rectangles were placed randomly and their size varied from 3 to 10 pixels.
    The color was chosen at random (with RGB components ranging from $0$ to $255$).
    The VAE loss target was set at $25$.

    \begin{itemize}
        \item {\bf Classifier architecture}: 
        $\oC(1, 1, 8) \rightarrow \oC(3, 2, 16) \rightarrow \oC(1, 1, 16) \rightarrow \oC(3,2,32) \rightarrow \oC(1,1,32) \rightarrow \oC(3,2,48) \rightarrow \oC(1,1,48) \rightarrow \oA \rightarrow \oFC(2)$ with $\RR$ nonlinearities.
        \item {\bf Mask architecture}: $[\oC(1, 1, 8) \rightarrow \oC(3, 2, 16) \rightarrow \oC(1, 1, 16) \rightarrow \oC(3,2,32)] \rightarrow [\oC_s(3, 1, 16) \rightarrow \oR(10) \rightarrow \oC_s(3, 1, 16) \rightarrow \oR(16) \rightarrow \oC_s(3, 1, 8) \rightarrow \oR(32) \rightarrow \oC_s(3, 1 8) \rightarrow \oC(1, 1, 1)]$ with the subnetwork in the first half using $\RR$ and the network in the second half using $\LR$.
        \item {\bf Encoder architecture}: $\oC(3, 2, 8) \rightarrow \oC(3, 2, 8) \rightarrow \oC(3, 2, 16) \rightarrow \oC(3,1,16)$ with $\LR$ nonlinearities.
        \item {\bf Decoder architecture}: $\oFC(64) \rightarrow \oFC(128) \rightarrow \oS(8\times 8) \rightarrow \oT_s(3, 2, 16) \rightarrow \oT_s(3, 1, 16) \rightarrow \oT_s(3, 2, 16) \rightarrow \oT_s(3, 1, 4)$ with $\LR$ nonlinearities.
    \end{itemize}

\subsection{Multiple {\tt MNIST} Digits}

    All synthetic multi-digit images were generated by placing 2 or 4 digits into the quadrants and randomly shifting them by at most 4 pixels.
    For the two- and four-digit datasets, the small digit was downsampled to $18\times 18$ and $14\times 14$ correspondingly.
    The target VAE loss was set at $50$.

    \begin{itemize}
        \item {\bf Classifier architecture}: 
        $\oC(1, 1, 4) \rightarrow \oC_s(3, 2, 4) \rightarrow \oC(1, 1, 8) \rightarrow \oC(3, 2, 8) \rightarrow \oC(1,1,16) \rightarrow \oC(3,2,16) \rightarrow \oC(1,1,16) \rightarrow \oC(3,2,16) \rightarrow \oA \rightarrow \oFC(10)$ with $\RR$ nonlinearities.
        \item {\bf Mask architecture}: $[\oC(1, 1, 4) \rightarrow \oC_s(3, 2, 4) \rightarrow \oC(1, 1, 8) \rightarrow \oC(3, 2, 8)] \rightarrow [\oR(12) \rightarrow \oC_s(3, 1, 16) \rightarrow \oP(1) \rightarrow \oR(28) \rightarrow \oC_s(3, 1, 16) \rightarrow \oR(56) \rightarrow \oC(1, 1, 16) \rightarrow \oC(1, 1, 1)$ with the subnetwork in the first half using $\RR$ and the network in the second half using $\LR$.
        \item {\bf Encoder architecture}: $\oC_s(3, 2, 16) \rightarrow \oC(3, 2, 16) \rightarrow \oC(3, 2, 16) \rightarrow \oC(3,1,8)$ with $\LR$ nonlinearities.
        \item {\bf Decoder architecture}: $\oFC(24) \rightarrow \oFC(49) \rightarrow \oS(7\times 7) \rightarrow \oT_s(3, 2, 16) \rightarrow \oT_s(3, 1, 16) \rightarrow \oT_s(3, 2, 8) \rightarrow \oT_s(3, 2, 4) \rightarrow \oC(1, 1, 2)$ with $\LR$ nonlinearities.
    \end{itemize}

\subsection{{\tt SVHN}}

    All {\tt SVHN} images were cropped and down- up-sampled to $128\times 128$.
    We used the Inception-based image augmentation technique leaving at least $95\%$ of the entire number bounding box within the frame.
    The image transformation was not permitted to generate a crop containing less than $40\%$ of the original image.
    The VAE loss target was chosen to be at $2000$ and $\sigma = (1/8)^{1/2}$.

    \begin{itemize}
        \item {\bf Classifier architecture}: 
        $\oC_s(3, 1, 4) \rightarrow \oC_s(3, 2, 4) \rightarrow \oC_s(3, 1, 4) \rightarrow \oC_s(3, 2, 4) \rightarrow \oC_s(3, 1, 4) \rightarrow \oC_s(3, 2, 8) \rightarrow \oC_s(3, 1, 8) \rightarrow \oC_s(3, 2, 8) \rightarrow \oC_s(3, 1, 8) \rightarrow \oC_s(3, 2, 8)$ with $\RR$ nonlinearities.
        \item {\bf Mask architecture}: $[\oC_s(3, 1, 4) \rightarrow \oC_s(3, 2, 4) \rightarrow \oC(1, 1, 8) \rightarrow \oC(3, 2, 8)] \rightarrow [\oC_s(3, 1, 8) \rightarrow \oC_s(5, 1, 8) \rightarrow \oR(16) \rightarrow \oC_s(5, 1, 8) \rightarrow \oC_s(5, 1, 8) \rightarrow \oR(32) \rightarrow \oC_s(3, 1, 4) \rightarrow (1, 1, 1) \rightarrow \oR(128)$ with the subnetwork in the first half using $\RR$ and the network in the second half using $\LR$.
        \item {\bf Encoder architecture}: $\oC_s(3, 2, 8) \rightarrow \oC_s(3, 1, 8) \rightarrow \oC_s(3, 2, 16) \rightarrow \oC_s(3, 1, 16) \rightarrow \oC_s(3, 2, 16) \rightarrow \oC_s(3, 1, 16) \rightarrow \oC_s(3, 2, 16) \rightarrow \oC_s(3, 2, 32) \rightarrow \oC_s(3, 1, 32)$ with $\LR$ nonlinearities.
        \item {\bf Decoder architecture}: $\oFC(64) \rightarrow \oFC(128) \rightarrow \oS(8\times 8) \rightarrow \oT_s(3, 2, 16) \rightarrow \oT_s(3, 1, 16) \rightarrow \oT_s(3, 2, 16) \rightarrow \oT_s(3, 1, 16) \rightarrow \oT_s(3, 2, 8) \rightarrow \oT_s(3, 1, 8) \rightarrow \oT_s(3, 2, 4)$ with $\LR$ nonlinearities.
    \end{itemize}

\section{Supplementary Figures}
\label{sec:figures}

    \begin{figure*}[htpb]
        \centering
        \begin{subfigure}{.49\textwidth}
          \centering
          \includegraphics[width=.99\linewidth]{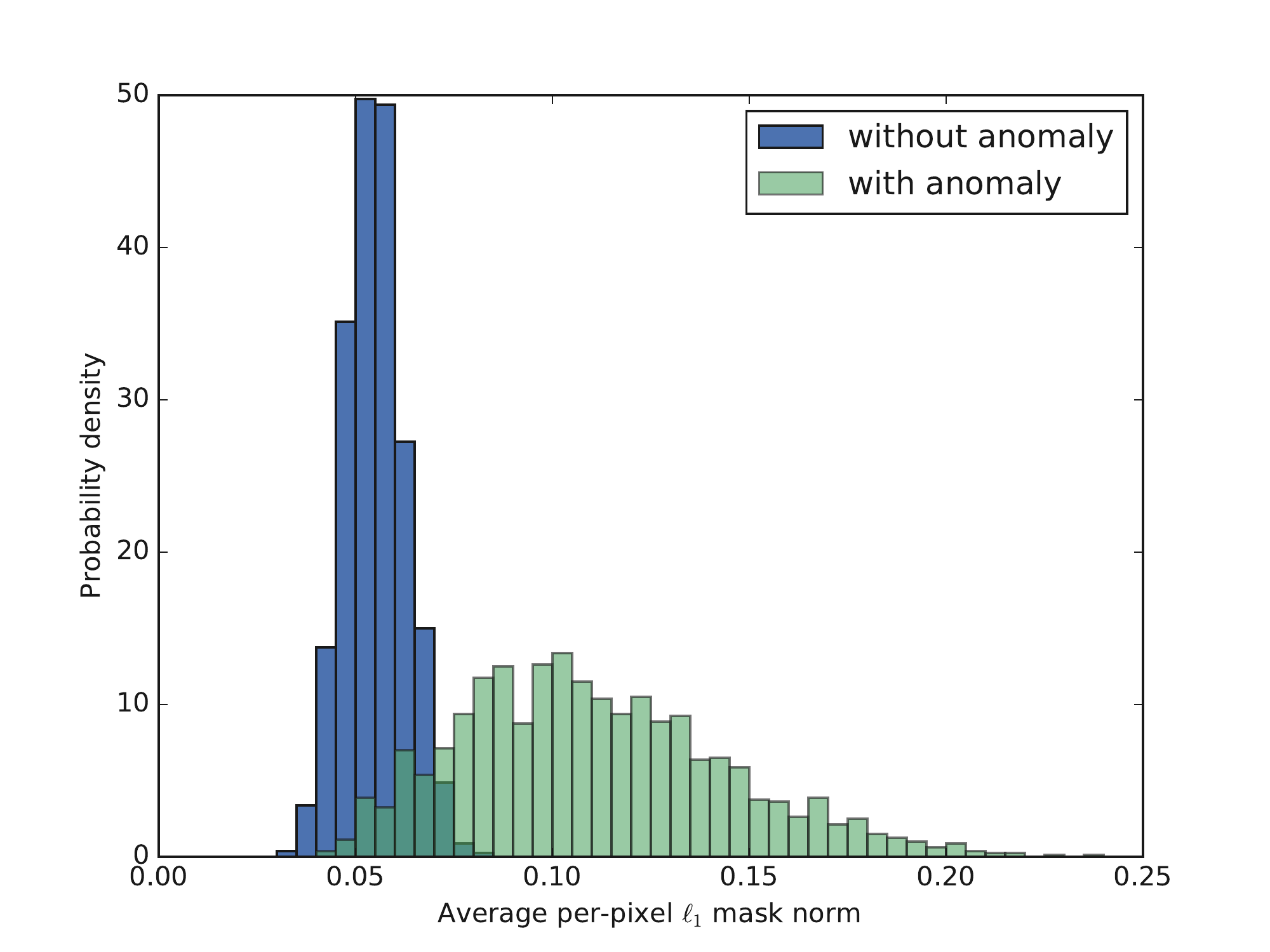}
          \caption{}
        \end{subfigure}
        \begin{subfigure}{.49\textwidth}
          \centering
          \includegraphics[width=.99\linewidth]{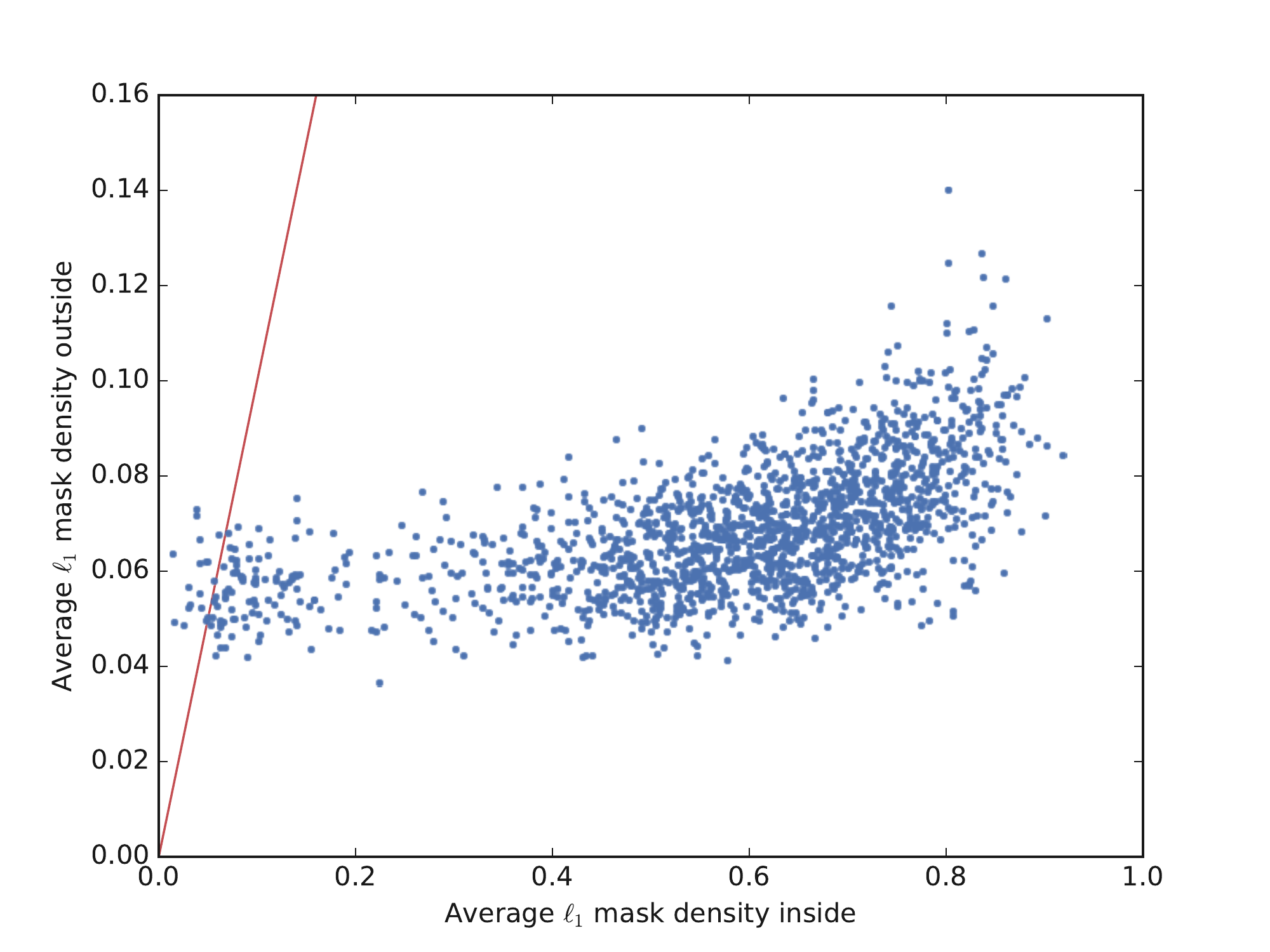}
          \caption{}
        \end{subfigure}
        \caption{Statistics of the mask model trained on {\tt MNIST} with rectangular patches: (a) histograms of the per-pixel average $\ell_1$ mask norms calculated for images with and without anomalies; (b) per-pixel average $\ell_1$ mask norm inside and outside of the rectangular patch for images with ``anomalies''.}
        \label{fig:mnist_stats}
    \end{figure*}

    \begin{figure*}[htpb]
        \centering
        \begin{subfigure}{.49\textwidth}
          \centering
          \includegraphics[width=.99\linewidth]{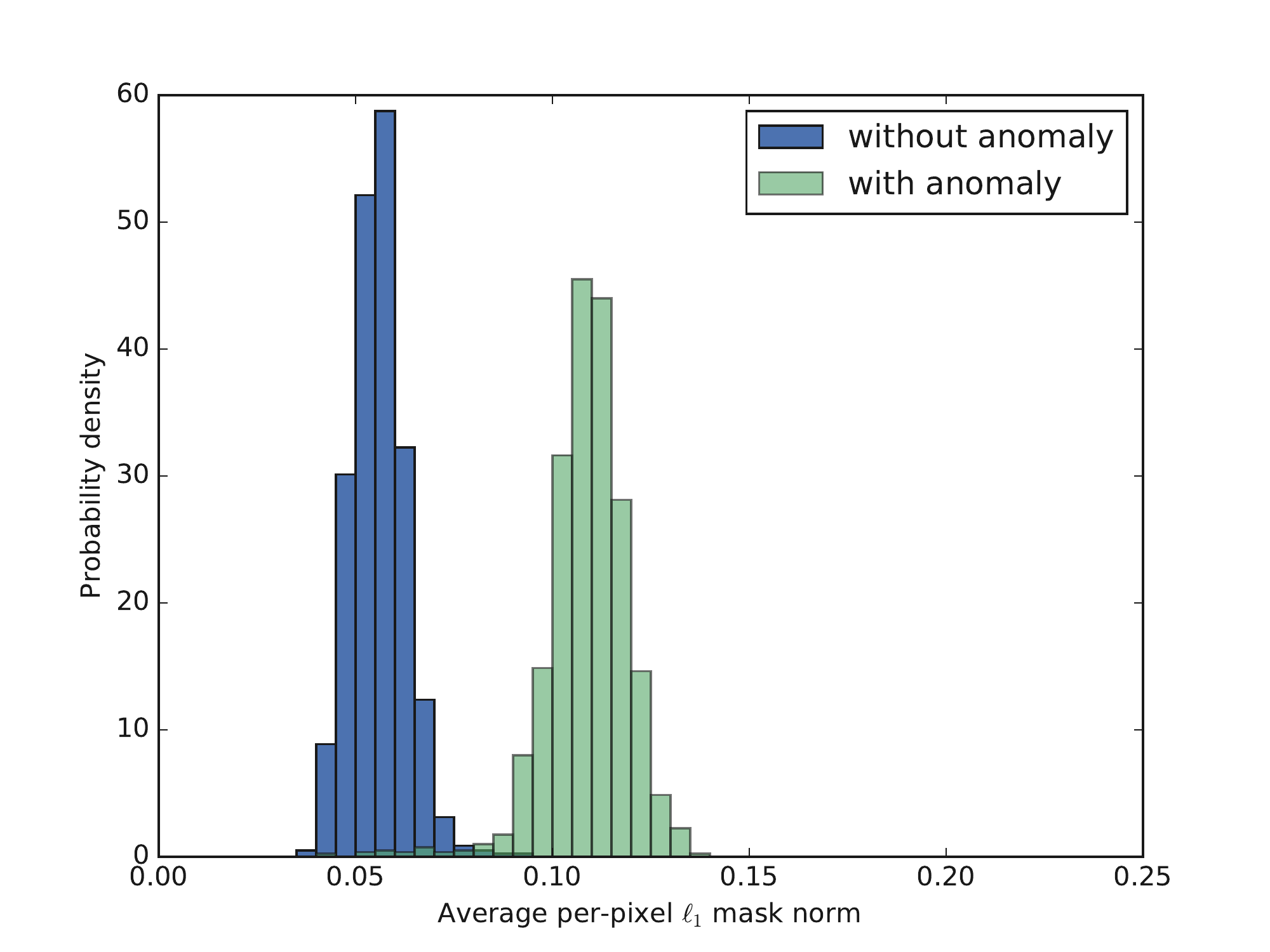}
          \caption{}
        \end{subfigure}
        \begin{subfigure}{.49\textwidth}
          \centering
          \includegraphics[width=.99\linewidth]{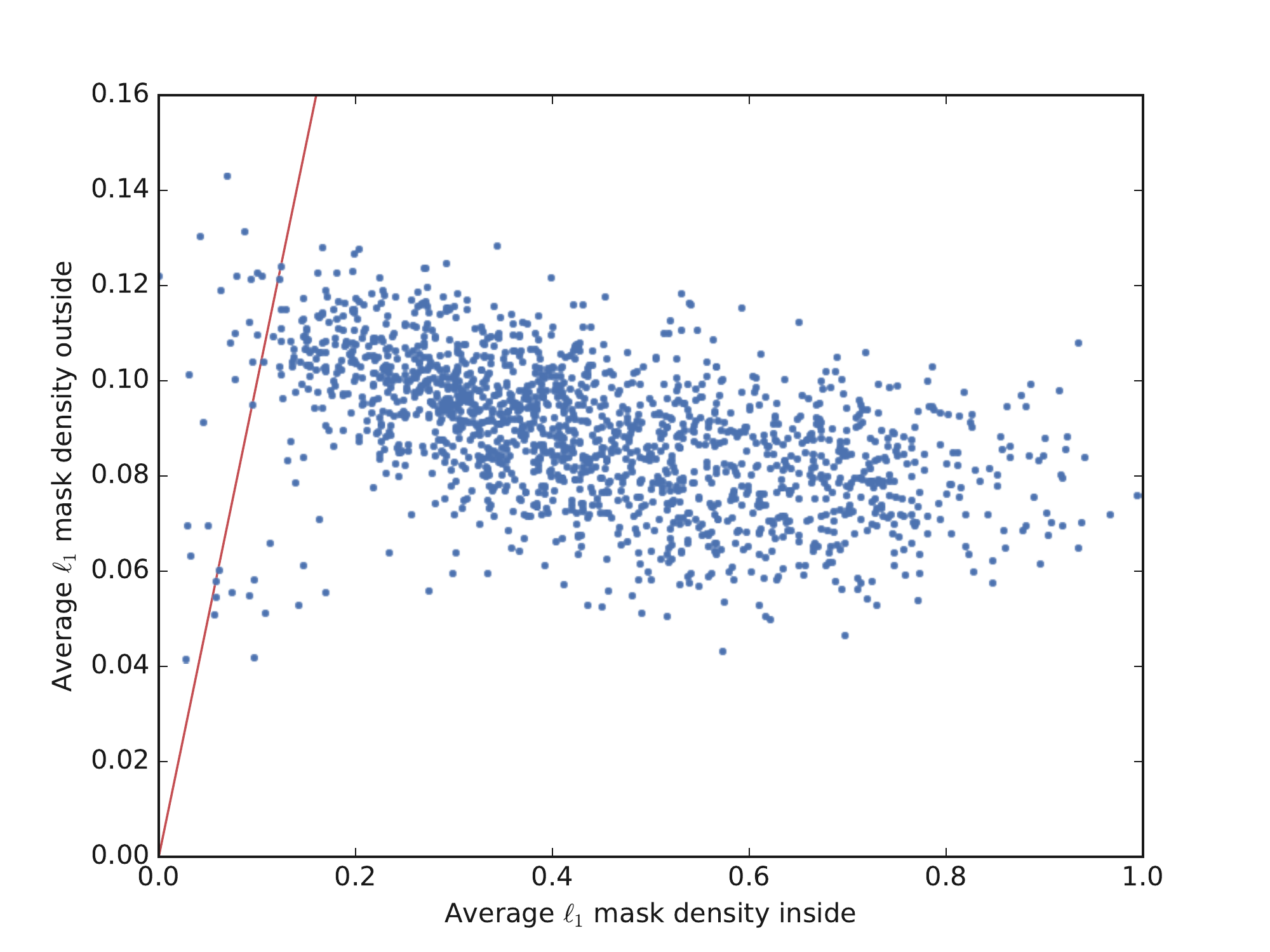}
          \caption{}
        \end{subfigure}
        \caption{Same as Figure~\ref{fig:mnist_stats}, but for {\tt CIFAR10} with rectangular patches. Unlike for the {\tt MNIST} dataset, the $\ell_1$ norm is seen to be a better predictor of the ``anomaly'' (see (a)), but produced masks appear to be less accurate (see (b)).}
        \label{fig:cifar_stats}
    \end{figure*}

    \begin{figure} 
        \centering
        \includegraphics[width=0.45\textwidth]{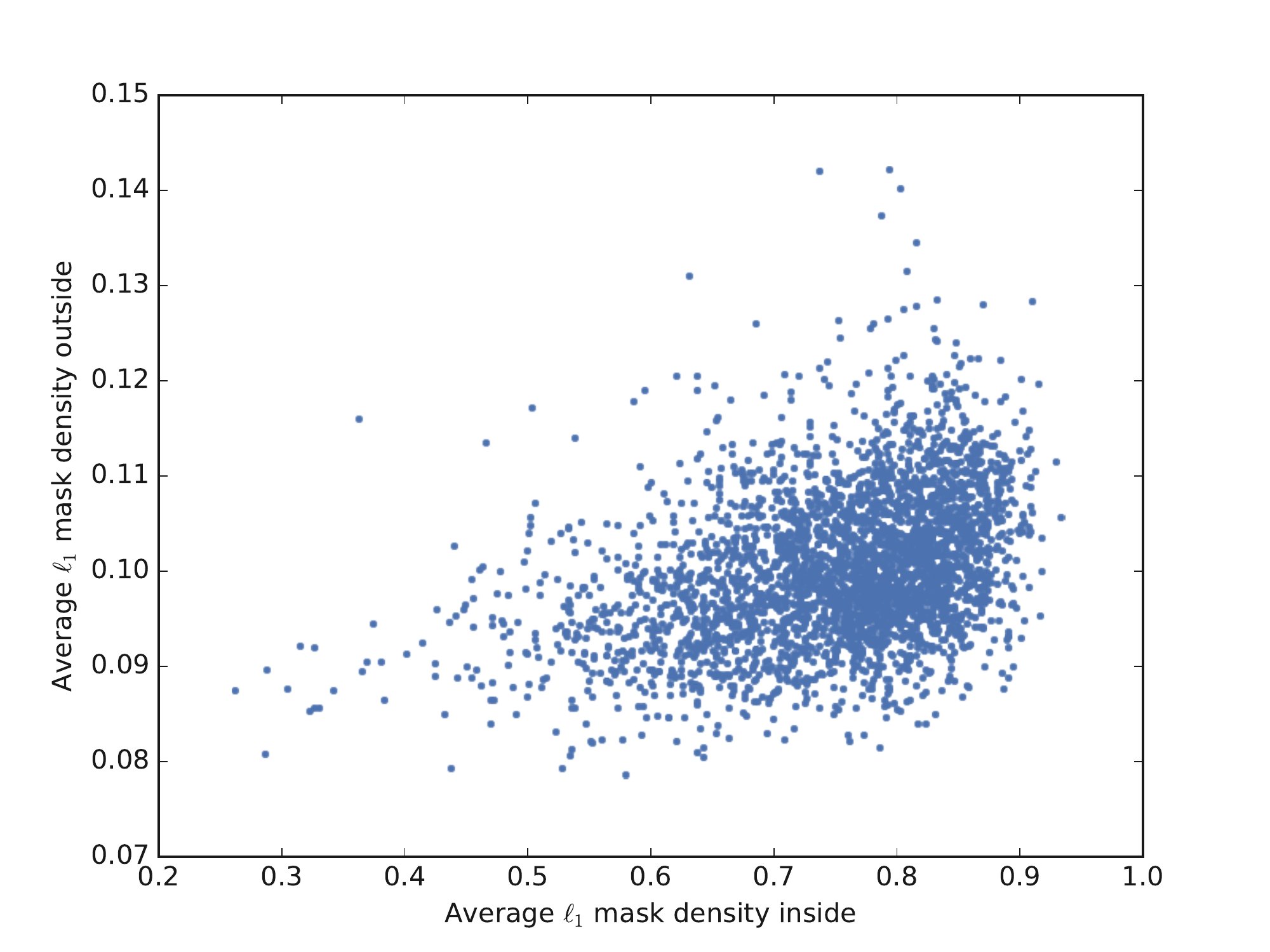}
        \caption{Average per-pixel $\ell_1$ mask norm outside and inside a circle of radius $8$ drawn around the location of the small digit (with the size $14\times 14$) in the {\tt MNIST}-based dataset with 4 digits.}
        \label{fig:quad_stats}
    \end{figure}

\end{document}